# Hierarchical entropy and domain interaction to understand the structure in an image

Author

Nao Uehara, Teruaki Hayashi, Yukio Ohsawa

Abstract

In this study, we devise a model that introduces two hierarchies into information entropy. The two hierarchies are the size of the region for which entropy is calculated and the size of the component that determines whether the structures in the image are integrated or not. And this model uses two indicators, hierarchical entropy and domain interaction. Both indicators increase or decrease due to the integration or fragmentation of the structure in the image.
It aims to help people interpret and explain what the structure in an image looks like from two indicators that change with the size of the region and the component. First, we conduct experiments using images and qualitatively evaluate how the two indicators change. Next, we explain the relationship with the hidden structure of Vermeer's girl with a pearl earring using the change of hierarchical entropy. Finally, we clarify the relationship between the change  of domain interaction and the appropriate segment result of the image by an experiment using a questionnaire.

KeyWord

Structual Change, Entropy, Hierarchy, Segmentation

1. Introduction

   We proposed hierarchical entropy as a system for humans to interpret and use the meaning of the structure in an image. In this system, a human interprets the change in the structure in an image using a newly proposed measure, known as hierarchically computed entropy. Humans interpret the meaning of a structure by establishing an explanatory relationship between different parts of an image. This particular characteristic differentiates human beings from machines. For example, this is different from the task that a machine accomplishes from learning; in this research, humans discover the meaning of the structure of an image from the computational results. The computational result, which is the hierarchical entropy, is the degree of integration of the structure.

   Entropy represents the clutter of a system. In this research, entropy is extended to represent the clutter of a system as a collection of each cluster formed by individual structures of an image in a certain region. Then, we changed the size of the region and the size of the clusters to capture the change in the structure. The size and number of structures contained in the region also differed after changing the size of the region in which the structure is captured. Smaller region sizes captured the parts of a dog's face (Figure 1). By observing the eyes, ears, and nose of a dog, we can understand its structure. It is possible to capture in more detail each structure in a single region that contains many structures. In addition, by changing the size of the clusters formed by each structure, the characteristics of the structures changed significantly. By changing the size of the clusters, the outline of a dog's face appeared. The eyes and nose were attached together to form a single structure (Figure 2). In addition, by changing "the scale of the region" and "the scale of the cluster," we assumed that the structure of the image would change. The changes are explained in terms of hierarchical entropy, which is an extension of information entropy used in the field of information science.

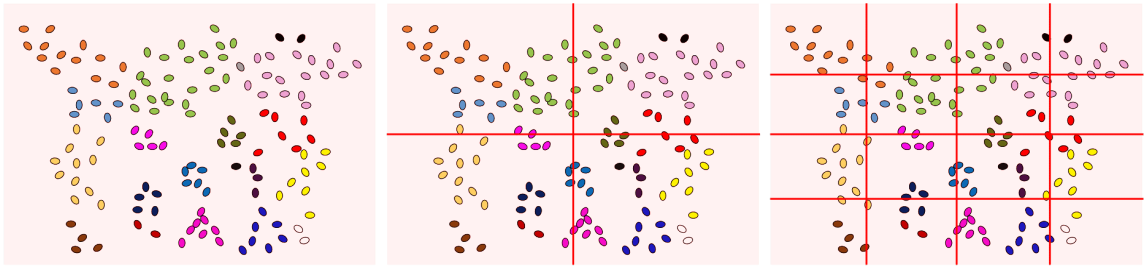

Figure 1. Make the region size smaller to understand the structure in more detail

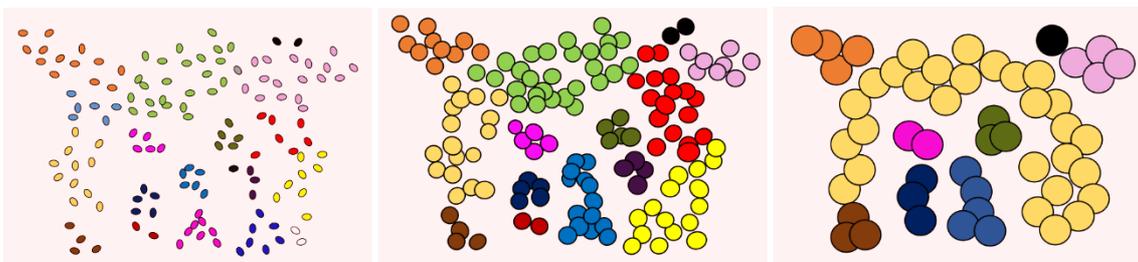

Figure 2. By changing the cluster size, the structure of the image also changes.

2. Related Research

The concept of entropy introduced in this study is discussed in information science. In thermodynamics, entropy can be defined as the energy divided by temperature as a physical parameter of the system[1]. In statistical mechanics, entropy is a measure quantified by the average logarithm of the number of possible microscopic arrangements of particles in a system, observed as a macroscopic disorder[2]. However, in the field of information science, entropy is defined as the amount of information required to specify the entire set of microscopic states of a system that is not specified by macroscopic variables. In this paper, entropy refers to the definition in information science, and the clusters defined below are considered as microstates. Entropy is then introduced here to quantify the diversity of clusters that belong to a region of a certain size.

We develop our method in the context of how information entropy has been applied in the past. First, the proposed method aims to compute the entropy based on clusters to capture changes in the structure within an image. It does not use entropy as a clustering criterion. This is similar to the marketing world where entropy has been used as an indicator of interest and product diversity [3, 4, 5]. In the field of image recognition, the temporal change in the entropy of each part of the image is used to detect contours and their movements [6, 7, 8, 9]. Information entropy has been widely used as an evaluation metric in texture analysis [10, 11], image restoration [12, 13], object detection, and 3D image capture [14, 15]. This technology has also been applied in fields like medicine and crime prevention. Furthermore, the entropy of traffic and events in computer networks has proven to be an important characteristic for detecting unexpected behavior and sudden changes. With this background, the aim of this paper is to introduce entropy into the field of image recognition as a measure of the change in the structures in an image, and to discover specific structures in an image based on the degree of change in entropy.

Graph-based entropy (GBE) [5] is an extension of information entropy [16] to graph structure, and it is proposed to detect signs of structural changes in data that are useful in explaining potential changes in consumer behavior. GBE captures the structural changes in the clusters formed by the graph. An example is presented in Figure 3. As seen in Figure 3 (a), there is only one large cluster, but in (b), one cluster is divided into five clusters. In (b), one cluster is split into five clusters. In this case, the entropy increases because the cluster is split. In contrast, the three clusters that exist in (c) are transformed into one large cluster in (d) because of the nodal point A. In this case, the entropy decreases because the clusters are integrated. Thus, the value of entropy changes with a change in the structure of the cluster. The nodal point A represents the discovery of an opportunity to increase product sales in POS data analysis, or the discovery of a precursor to seismic activity in seismic data [17, 18].

In the earthquake data, the precursors of earthquakes are determined by calculating the entropy change of the earthquake clusters formed in a region of predetermined size [19]. In this study, the size of the region was fixed, while the size of the clusters changed over time. We now introduce the idea of a region scale for calculating entropy into GBE which we further extended by introducing the scale of the clusters.

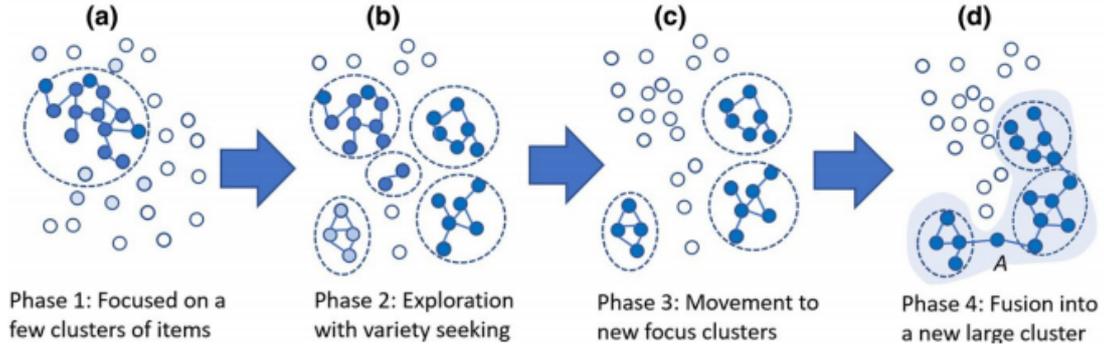

Figure 3. As the graph structure transitions from (a), (b), (c), and (d), the cluster structure also changes (except from paper [5]).

3. Proposed Method (Hierarchical entropy)

3.1 Definition of Hierarchy

In graph-based entropy, the change in entropy for time series data is used to explain the change in sales of products. However, we define hierarchy by noting that image data are not time-series data in this research. We introduce two types of hierarchies: the hierarchy of the region and the hierarchy of the cluster.

First, we define this region. The region was determined by layer $\mathcal{L}$ (Figure 4). An image is divided into several regions of arbitrary scale size $S$ in layer $\mathcal{L}$. As the size of each region reduces, the number of regions increases. Each rectangular region represents one region (Figure 1). We define layer $\mathcal{L}_n$ as the size of the $n$-th scale. Let $S_i^{\mathcal{L}_n}$ denote the $i$-th size region $S$ in layer $\mathcal{L}_n$. The hierarchical entropy in this region is given by $S_i^{\mathcal{L}_n}$. Here, $\mathcal{L}_n$ is the upper region and $\mathcal{L}_{n+1}$ is the lower region. The hierarchy of the region represents each layer. By changing the region, it is possible to observe and compute locally or globally. The finer the region, the more fragmented is the structure in the image. Changing the hierarchy of the region causes the loss and generation of structures. By examining the change in the hierarchical entropy, it is possible to explain the disappearance and occurrence of structures in the region.

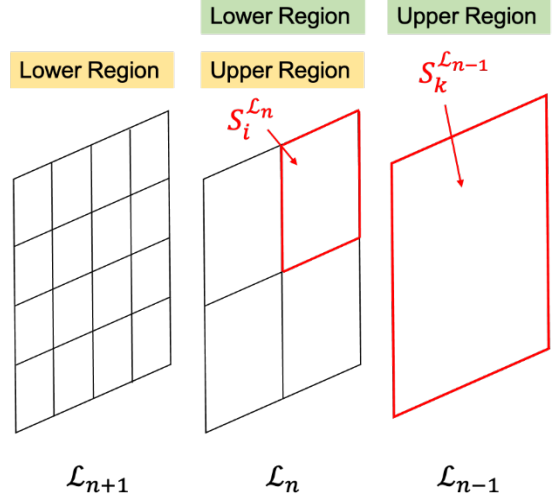

Figure 4. Relationship between Layer $\mathcal{L}$ and region $S$

Next, we define a cluster. A cluster is a collection of pixels with similar RGB values, defined by using efficient graph-based image segmentation (EGBIS) [20]. The parameter that determines whether RGB is similar is the component size $|C|$, and the total number of clusters is determined by the component size $|C|$. The larger the component size $|C|$, the more likely it is for the pixels to merge together, and larger the cluster size. Furthermore, the hierarchy of a cluster represents the degree of component size.

3.2 Definition of Entropy and Domain Interaction

After introducing the above two hierarchies, we now define the hierarchical entropy. The hierarchical entropy $H(S_i^{\mathcal{L}n})$ is the entropy of the probability $P_i^{\mathcal{L}n}(c_k)$ of the existence of a group of clusters in a region $S_i^{\mathcal{L}n}$. The parameter $c_k$ is the $k$-th cluster among all the clusters determined by the component size $|C|$. In this case, the hierarchical entropy of region $S_i^{\mathcal{L}n}$ is $H(S_i^{\mathcal{L}n})$ defined by equation (1). $P_i^{\mathcal{L}n}(c_k)$ in Equation (1) is defined in Equation (2). $|S_i^{\mathcal{L}n}|$ in Equation (2) is the total number of pixels contained in region $S_i^{\mathcal{L}n}$. Thus, $P_i^{\mathcal{L}n}(c_k)$ is the number of datapoints belonging to cluster $c_k$ divided by the total number of datapoints of $|S_i^{\mathcal{L}n}|$ in region $S_i^{\mathcal{L}n}$.

$$H(S_i^{\mathcal{L}n}) = -\sum P_i^{\mathcal{L}n}(c_k) \log P_i^{\mathcal{L}n}(c_k). \qquad (1)$$

$$P_i^{\mathcal{L}n}(c_k) = \frac{1}{|S_i^{\mathcal{L}n}|} \sum_{x_j \in S_i^{\mathcal{L}n}} \delta_{j,k}. \qquad (2)$$

$$\delta_{j,k} = \begin{cases} 1 & (x_j \in c_k) \\ 0 & (x_j \notin c_k) \end{cases}. \qquad (3)$$

As described above, we can calculate the entropy $H(S_i^{\mathcal{L}_n})$. In addition, we explain the layers introduced in this study. Layer L has an upper layer $\mathcal{L}_{n-1}$ and a lower region $\mathcal{L}_n$, where there is a hierarchy between each layer (Figure 1). Equation (4) shows that the data set $\mathcal{X}$ is divided in each region $S$.

$$\mathcal{X} = \bigcup_{S_j^{\mathcal{L}_n} \in \mathcal{L}_n} \{x_j | x_j \in S_j^{\mathcal{L}_n}\} = \bigcup_{S_k^{\mathcal{L}_{n-1}} \in \mathcal{L}_{n-1}} \{x_k | x_k \in S_k^{\mathcal{L}_{n-1}}\}. \qquad (4)$$

As shown in Equation (4), by viewing the dataset at different scales, large and small, we can consider a global view of the dataset instead of a local view. This allowed us to analyze the data in a cross-sectional manner. In addition, several regions $S^{\mathcal{L}_n}$ in the same layer $\mathcal{L}_n$ should not overlap. In other words, Equation (5) holds for two regions $S_i^{\mathcal{L}_n}$ and $S_j^{\mathcal{L}_n}$ in the same layer $\mathcal{L}_n$ that are disjoint. Datasets that overlap in the two regions $S_i^{\mathcal{L}_n}$ and $S_j^{\mathcal{L}_n}$ become empty sets.

$$\bigcup \{x_i | x_i \in S_i^{\mathcal{L}_n} \land x_i \in S_j^{\mathcal{L}_n}\} = \emptyset. \qquad (5)$$

Thus, we defined the partitioned computational region $S$ and $H(S_i^{\mathcal{L}_n})$.

We also defined domain interaction as an indicator used in this study to describe the changes caused by integrating clusters in Equation (6). The domain interaction is defined as $I(S_i^{\mathcal{L}_n})$, and it represents the interaction between the upper region $S_i^{\mathcal{L}_n}$ and the lower region $S_j^{\mathcal{L}_{n+1}} \subset S_i^{\mathcal{L}_n}$ (Figure 5). $S_i^{\mathcal{L}_n}$ consists of several lower regions $S_j^{\mathcal{L}_{n+1}}$. As seen in Figure 5, one $S_i^{\mathcal{L}_n}$ consists of four lower regions $S_j^{\mathcal{L}_{n+1}}$. $\mathcal{B}_k^{\mathcal{L}_n}$ is a set of clusters that exist in the upper region $S_i^{\mathcal{L}_n}$ and straddle the lower regions $S_j^{\mathcal{L}_{n+1}}$. The reason $I(S_i^{\mathcal{L}_n})$ can be called the interaction between regions is as follows: $I(S_i^{\mathcal{L}_n})$ can be interpreted as a penalty for splitting the clusters in the upper region $S_i^{\mathcal{L}_n}$ into the lower region $S_j^{\mathcal{L}_{n+1}} \subset S_i^{\mathcal{L}_n}$. When many clusters in different regions $S_j^{\mathcal{L}_{n+1}}$ merge into a single cluster, the domain interaction becomes larger. In other words, it is possible to understand the integration or fragmentation of clusters within a region $S_i^{\mathcal{L}_n}$ from changes in domain interaction.

$$I(S_i^{\mathcal{L}_n}) = \sum_{c_l \in \mathcal{B}_k^{\mathcal{L}_n}} P(c_l) \sum_{S_j^{\mathcal{L}_{n+1}} \subset S_i^{\mathcal{L}_n}} \{-P(S_j^{\mathcal{L}_{n+1}} | c_l) \log P(S_j^{\mathcal{L}_{n+1}} | c_l). \qquad (6)$$

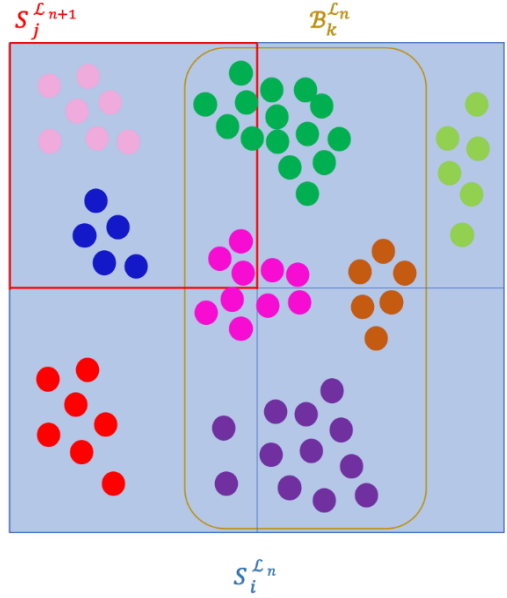

Figure 5. The concept of $\mathcal{B}_k^{\mathcal{L}n}$

In this study, we used two metrics to describe the change in structure: hierarchical entropy and domain interaction. Hierarchical entropy is defined as $H(S_i^{\mathcal{L}n})$ and domain interaction is defined as $I(S_i^{\mathcal{L}n})$. When $H(S_i^{\mathcal{L}n})$ is large, it means that the state of the system in the region $S_i^{\mathcal{L}n}$ is cluttered and has a high degree of dissipation. Meanwhile, when $H(S_i^{\mathcal{L}n})$ is small, the state of the system is biased towards one state, which means that the dissipation is low.

The novelty of this research is that it is possible to combine a model that extends information entropy and introduces hierarchy with EGBIS to help people explain how the structure of the image changes after a change in the hierarchical entropy and domain interaction. This is different from recent tasks of improving the accuracy of image recognition using computers and models. It is an image recognition task that can explain the calculated results to humans.

4 Results

4.1 Data and parameters

In this study, artworks were randomly selected from WikiArt[1]. Artwork from WikiArt has also been used in research on clustering artworks [21, 22]. Here, we also compared the entropy and domain interactions of artworks of various styles and ages. All images used as data were converted into JPEG. The color space represents all the RGB values. Here, we did not change the size of the data. All images were smoothed using a Gaussian filter (σ=2.0) to remove noise.

---

[1] https://www.wikiart.org/

To derive the hierarchical entropy from the definition, we use the EGBIS algorithm to perform graph-based clustering of images. The size of each cluster was determined by component size. In this study, we used 19 variables of $|C| = \{100, 200, 300, 400, 500, 600, 700, 800, 900, 1000, 2000, 3000, 4000, 5000, 6000, 7000, 8000, 9000, 10000\}$ to perform cluster segmentation. The number of clusters$(= \mathcal{C})$ is determined by the component size, and $k$ of $c_k$ is expressed as $\{k = 1, 2, \ldots, |\mathcal{C}|\}$. The height of the image is represented by $H$ and the width is represented by $W$. The computational region is determined by layer $\mathcal{L}$. Layers $\mathcal{L}$ is set to $\mathcal{L}_n$ for each image segment. Here, we assume that $n$ varies from 0 to 6 and the image is divided into $2^n \times 2^n$ segments. Entropy was calculated under the above conditions.

The domain interaction varies depending on the size and number of clusters that cross the lower regions. The domain interactions were calculated for each upper region where each region was divided into $2 \times 2$ segments. From the change in the value of the domain interaction within each $S_i^{\mathcal{L}_n}$, the change in the structure of the clusters was verified.

4.2 Experimental Results

In this section, we discuss the results of entropy calculations. First, we consider the results of the change in the entropy value when the region size $S$ changes, and then, we discuss the change when the component size is varied. We evaluated the results qualitatively.

4.2.1 Change in entropy as region size is varied

We discuss the change in the entropy $H(S^{\mathcal{L}_n})$ of region $S$ when layer $\mathcal{L}_{n \{n|n=0,1,2,3,4,5,6\}}$ is changed. Here, the component size is fixed. The entropy $H(S^{\mathcal{L}_n})$ of each region $S_i^{\mathcal{L}_n}$ is the average entropy $H(S_i^{\mathcal{L}_n})$ (Equation 7). We hypothesize that the smaller the size of the region, the more detailed is the structure that could be captured.

$$H(S^{\mathcal{L}_n}) = average\big(H(S_i^{\mathcal{L}_n})|i = 1, 2, \ldots, 2^n \times 2^n\big). \tag{7}$$

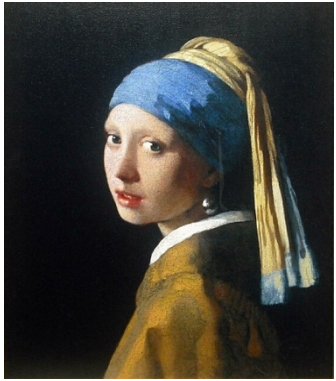

Figure 6. Images used to verify the entropy change.

We discuss the change in entropy for each component size Figure 7 (a). First, for each component size, the entropy $H(S^{L_n})$ decreases as the region becomes finer. This is because the finer the region, the smaller the resolution at which the clusters are captured, which leads to more regions being occupied by a single cluster. In other words, as $n$ increases, the number of regions with entropy of 0.0 increases, and the average entropy $H(S^{L_n})$ decreases.

Consider the hierarchy among the regions. The image in Figure 7(b), for which entropy is calculated using a 4×4 split, has a higher entropy at its center. This could mean that there are more clusters at the center of the image. A clustered image indicates that the image contains many clusters, which means that the image has a relatively complex structure around it. Meanwhile, the finer the region size, the more finely we can determine which parts of the image have high entropy values. In particular, when the image is divided into 64×64 segments, it is possible to detect the borderline areas of the clusters in the image. The cluster boundaries indicate that multiple clusters are mixed in the image. The hierarchy of region sizes allows us to capture structures both globally and locally. In a previous study [23], information entropy was used to detect edges more clearly. From this study, it can be concluded that smaller the size of the region, the easier it is to capture the edges between clusters. Meanwhile, global capturing makes it easier to understand where information (clusters) are concentrated in an image.

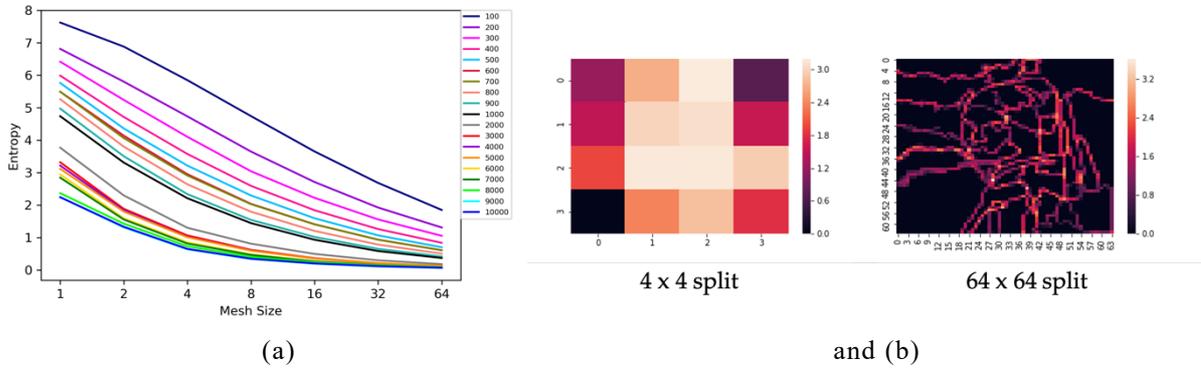

(a) and (b)

Figure 7. The change of entropy (a) Relationship between region size and entropy (b) Entropy of each region (component size is 1000)

4.2.2 Change in entropy when component size is changed

We fixed the size of the region and considered the variation in entropy $H(S^{L_n})$ with the component size. As the component size increased, the clusters were integrated, and the information possessed by the image diminished. For this reason, we assumed that the region with high entropy when the component size is sufficiently large is the part with more information.

Figure 9 (a) shows the graph of entropy change with component size from 100 to 1000, and (b) shows the graph of entropy change with component size from 1000 to 10000. The entropy $H(S^{L_n})$ decreases as the component size increases. This is because as the component size increases, the total number of clusters decreases, and the clusters integrate with each other. This suggests that the information possessed by the image is concentrated. This can be observed from the fact that an image with a component size of 100 is a mosaic while an image with a component size of 1000 captures the outline and facial features of a girl. In other words, the miscellaneous clusters are concentrated in the girls' clusters. However, when the component size is 10000, only the outline of the girl is captured. This means that the detailed information of the girl (eyes, nose, and mouth) is also available. Compared with the original image, we can distinguish the girls' features from the background of the structure. The rate of change (slope in Figure 8) of the entropy indicates the degree of integration of the clusters. We can assume that there is a large structural change before and after the component size, which caused the entropy to decrease sharply.

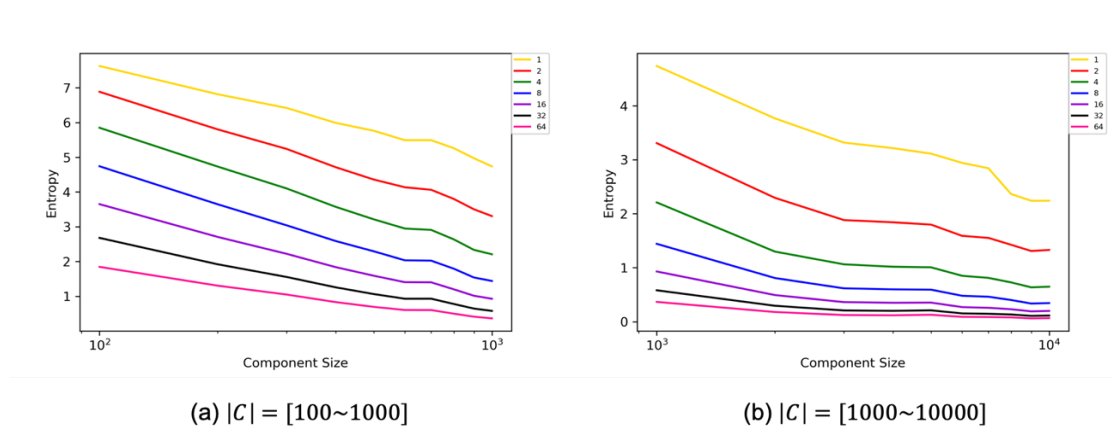

(a) $|C| = [100\sim1000]$     (b) $|C| = [1000\sim10000]$

Figure 8. Graph of the relationship between component size and entropy.

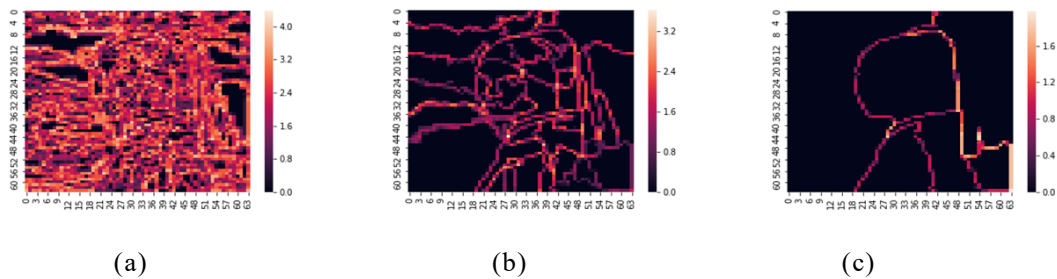

(a)                (b)                (c)

Figure 9. Entropy of each region (region size is 64×64 split) (a) component size is 100 (b) component size is 1000 (c) component size is 10000

   Here is an example of how the change in entropy led to the discovery of a structure. In Figure 10 (a), the area enclosed in blue is black, and there is no structure of particular interest. However, the blue areas in Figure 10 (b) and (c) are areas of high entropy, which can be interpreted as areas that break up the cluster. Even if the component size increased from 3000 to 7000, the entropy did not reach zero. This means that the lack of cluster integration can be explained by the presence of specific structures. Comparing images (a), (b), and (c) in Figure 10, we can predict that there are structures that cannot be captured by the naked eye. Therefore, we surveyed studies on the painting of a girl with pearl earring.

The original image of the girl with the pearl earring was scientifically investigated using non-invasive imaging and scanning techniques, digital microscopy, and paint sample analysis to create the original image (Figure 6) [24]. The results revealed the presence of a green curtain on an empty background (Figure 11). This is considered to correspond to the area enclosed in blue in Figure 10 (b) and (c). In this way, there are examples where the part where the entropy does not change can lead to the discovery of structures that are difficult for people to notice. The part with high entropy, even though the component size is large, can be interpreted as a specific structure where cluster integration does not proceed.

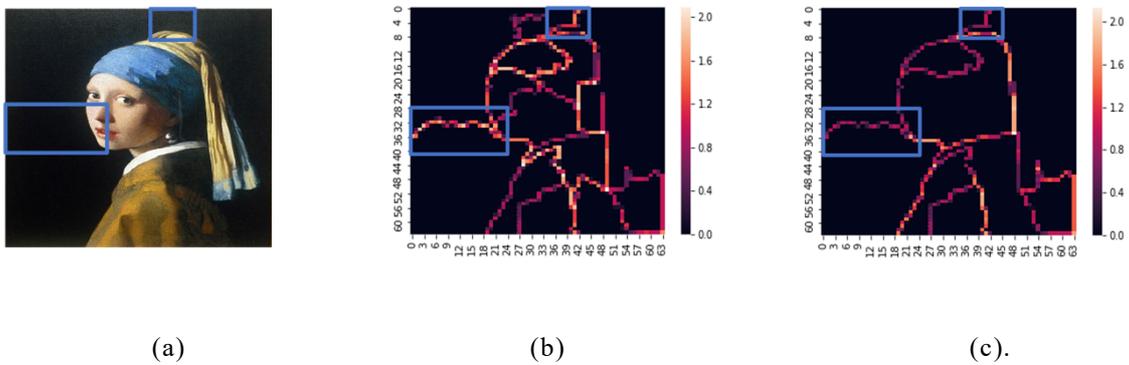

(a) (b) (c).

Figure 10. The relationship between entropy and structure identification (a) Original image (b) The entropy of each region (component size is 3000) (c) The entropy of each region (component size is 7000)

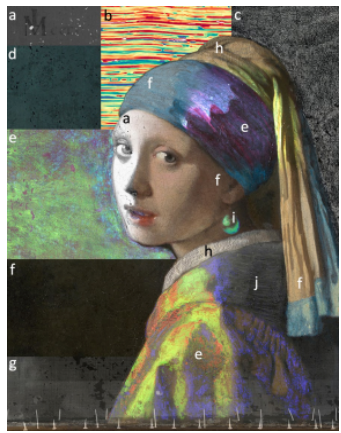

Figure 11. Painting of girl with a pearl earring created by scientific research (excerpt from [24])

4.3 Results of Domain Interaction Calculations

The size of the regions was fixed for calculations. As the region interaction increases with the increase in cluster integration, we assume that the region interaction increases monotonically with an increase in the component size. Further, we assumed that the slope would indicate the degree of integration.

4.3.1 Qualitative evaluation of domain interaction

Both (b) and (c) of Figure 8 show the change in the value of the domain interaction for this image (Figure 12 (a)). In both (b) and (c) of Figure 12, the vertical axis is the value of the domain interaction, and the horizontal axis is the component size that determines the size of the cluster. $[S_0, S_1, S_2, S_3]$ of Figure 9 follows $[S_0, S_1, S_2, S_3]$ in the image.

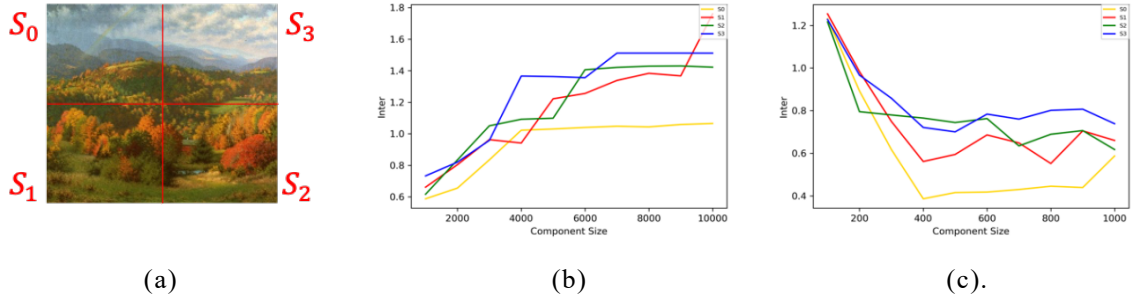

(a) (b) (c).

Figure 12. Relationship between domain interaction and component size (a) Image used to calculate domain interaction (b) The change of domain interaction (component size is from 1000 to 10000) (c) The change of domain interaction (component size is from 100 to 1000)

As shown in Figure 12 (b), as the component size increases in any of the four regions $[S_0, S_1, S_2, S_3]$, the value of the domain interaction also increases. This change explains the progress of cluster integration in any domain. This can be understood from the fact that as the component size increases, multiple clusters are integrated into larger clusters (Figure 13). The convergence of the domain interaction value to a certain value means that the clusters will not be integrated from there on. This is consistent with our assumption that domain interactions also increase as the component size increases.

However, as shown in Figure 12 (c), the domain interaction is different from the change in (b). For component sizes of 100 and 400, the domain interaction decreased. Then, between the component sizes of 400 and 1000, it fluctuated repeatedly. These changes are inconsistent with our hypothesis that domain interactions increase monotonically. As shown in Figure 13, the domain interaction should increase because the integration itself seems to be progressing. We believe that this change occurs for the following reasons: -

When the component size was 100, the size of each cluster across the lower domains was very small, and their number was small. Each of these clusters has a significant impact on the domain interaction. Therefore, domain interaction was high. Later, when the component size increased to 400, the domain interaction decreased because the size of the clusters across the domain was larger and the number of clusters increased because of the integration of clusters. When the component size was between 400 and 1000, there was no significant change in the domain interaction because the clusters were not integrated. Thus, the domain interaction changes differently from our hypothesis of a monotonic increase.

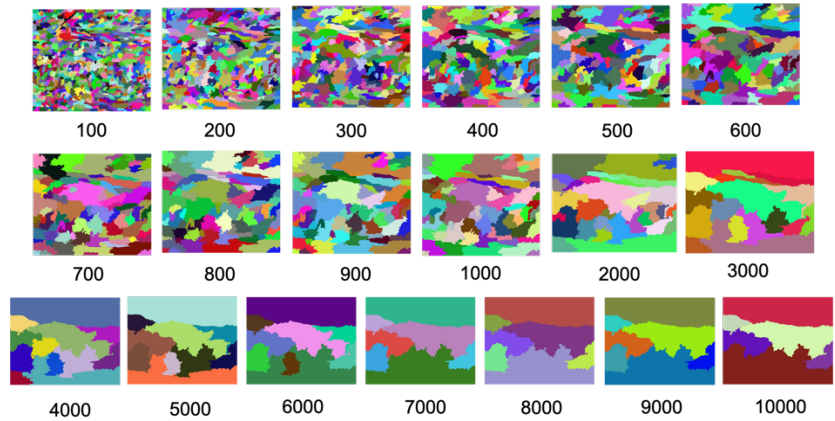

Figure 13. Image (Figure 8(a)) of region segmentation by component size from 100 to 10000

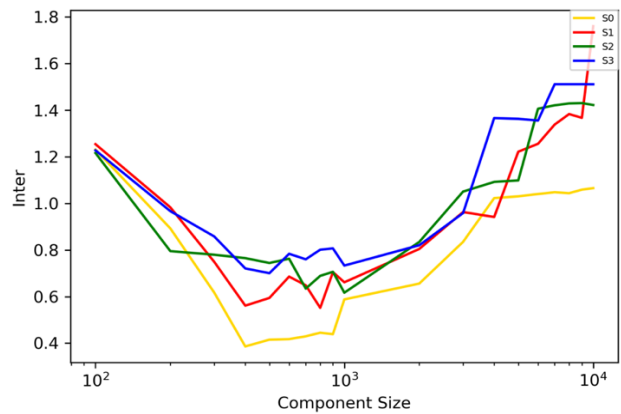

Figure 14. Variation of domain interaction (component size is from 100 to 10000)

4.3.2 Questionnaire for segment results

From the previous section, we found that the domain interaction decreases and then increases; therefore, we considered the meaning of the minimum value of the domain interaction.

When the component size increases from 100 to 10,000, the cluster goes from being a fine mosaic to a certain size, and then consolidates until there is no change. Taking Figure 13 as an example, the mosaic becomes a cluster of trees, clouds, and sky, and then the trees merge with each other and the sky with clouds to form a large cluster. The mosaic clusters become meaningful for people when the domain interaction is decreasing. Further, while the domain interaction is increasing, more clusters than necessary are being integrated. This means that fragmented clusters that need to be integrated are gradually integrated, and clusters that do not need to be integrated are integrated. In other words, we observed that the best balance between integration and fragmentation was achieved by dividing the clusters when the domain interaction was minimized. In this section, we test this assumption through an experiment using a questionnaire.

We considered the state where the balance of integration of clusters is optimal to be the image where the segment result is most appropriate for humans. We then tested whether the component size that minimized the domain interaction matched the component size used by humans when they considered the image to be the state with the most appropriate segment result. The experimental method was as follows: -

The eight component sizes were [200, 500, 800, 1000, 3000, 5000, 7000, 10000]. From the eight segmented images, the subjects were asked to select the image that they thought was correctly segmented with respect to the original image under the condition that multiple answers were allowed. Since the personal attributes of the respondents were not relevant, we did not collect any personal information from the respondents. The questionnaire was answered online using Google Forms. The sample size was 14. A total of 11 images were prepared.

The first step was to extract images that would result in the correct segmentation. In the questionnaire, we defined the correctly segmented images as those for which the response rate of the original image was 60% or more. As shown in Figure 15, the images were divided into two groups: Group (a) which was not segmented correctly, and Group (b) which was segmented correctly (Figure 15).

Next, we extracted the component size that resulted in the most correct segment for each image that resulted in correct segmentation. The numbers for each choice in Table 1 are the values of the component sizes used in the segment. The component size of the segmented image with the highest response rate was considered to be the correct component size (Table 1). This was then defined as $|C|'$. Then, the component size that minimized the domain interaction of each image in group B was defined as $|C|$. We verified whether this $|C|$ and $|C|'$ were consistent.

Table 1. Percentage of respondents who answered that the eight segment results in the image were correct (in %)

| | 200 | 500 | 800 | 1000 | 3000 | 5000 | 7000 | 10000 |
|---|---|---|---|---|---|---|---|---|
| 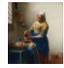 | 7.1 | 21.4 | 28.6 | 28.6 | 35.7 | 21.4 | 7.1 | 7.1 |
| 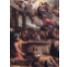 | 14.3 | 35.7 | 42.9 | 42.9 | 0 | 0 | 0 | 0 |
| 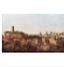 | 7.1 | 21.4 | 21.4 | 50 | 50 | 21.4 | 14.3 | 7.1 |
| 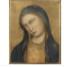 | 7.1 | 21.4 | 42.9 | **64.3** | 28.6 | 21.4 | 28.6 | 35.7 |
| 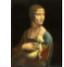 | 0 | 42.9 | **78.6** | 57.1 | 21.4 | 0 | 0 | 0 |
| 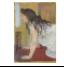 | 0 | 28.6 | **85.7** | 64.3 | 21.4 | 21.4 | 7.1 | 7.1 |
| 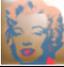 | 7.1 | 28.6 | 42.9 | **78.6** | 35.7 | 50 | 0 | 0 |
| 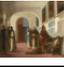 | 42.9 | 35.7 | **71.4** | 50 | 7.1 | 14.3 | 7.1 | 0 |
| 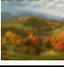 | 0 | 7.1 | 28.6 | 21.4 | 50 | **71.4** | 7.1 | 7.1 |
| 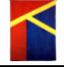 | 0 | 7.1 | 7.1 | 7.1 | 28.6 | 35.7 | **85.7** | 64.3 |
| 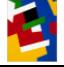 | 0 | 7.1 | 7.1 | 0 | 35.7 | **78.6** | 71.4 | 28.6 |

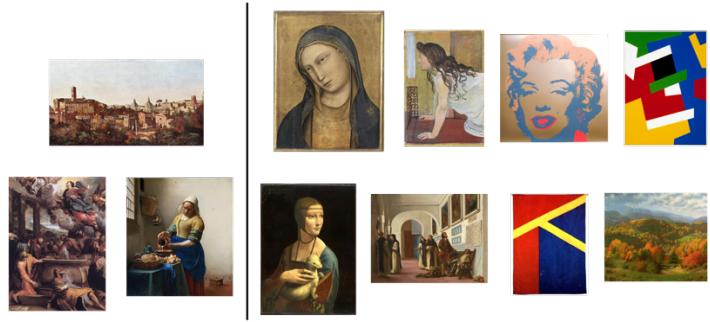

(a)            and (b)

Figure 15. Images classified by questionnaire results (a) A group of images for which the segment result was determined to be incorrect (b) A group of images for which the segment result was determined to be correct.

5.3.4 Relevance to Domain Interaction

We define $diff|C|$ as the difference between $|C|$ and $|C|'$ in Equation (7).

$$diff|C| = \big||C| - |C|'\big| \qquad (7)$$

Let group (a) be the images of $diff|C| < 500$ and group (b) be the images with which $diff|C| > 500$ (Figure 16). In group A, we considered that $|C|$ and $|C|'$ were congruent. The curves of the change in the values of domain interaction for groups (a) and

(b) of Figure 16 are shown in Figure 17 (a) and (b), respectively. The blue vertical line in the graph represents the component size, which was determined to be the correct segmentation results in the questionnaire. The green vertical line represents the component size at which domain interaction is minimized.

A possible reason for the disagreement between $|C|$ and $|C|'$ in group (b) is that the segmentation of images that were mostly occupied by a single color required a larger component size. Image (6) was dominated by red, blue, yellow, and green; image (7) was dominated by red and blue; and image (8) was dominated by green. If we did not increase the component size, these clusters, which should not be split, would be split, and we would not get the correct segmentation results. In group (a), on the other hand, the image was nicely segmented. The areas occupied by a single color were shaded or graded. As a result, we assumed that the segments were correct.

In other words, the images in group (a) were figurative paintings, such as people, whereas the images in group (b) were abstract paintings. This could mean that the two component sizes ($|C|$ and $|C|'$) were consistent in figurative paintings where the meaning was easier for people to understand. Figurative paintings were more likely to be divided into meaningful clusters when segmented because they had more details in terms of color and composition. We thought that for figurative images, it was possible to use the component size that minimized domain interaction to infer the component size that could be corrected for segmentation using EGBIS.

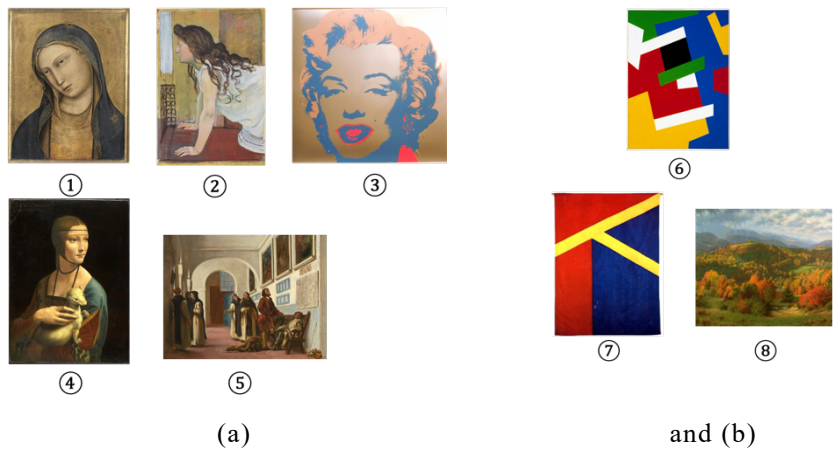

(a)  and (b)

Figure 16. Groups determined by the correct segment result (a) The component size that results in the minimum value of domain interaction matches one of the questionnaire (b) The component size that results in the minimum value of the domain interaction does not match one of the questionnaire.

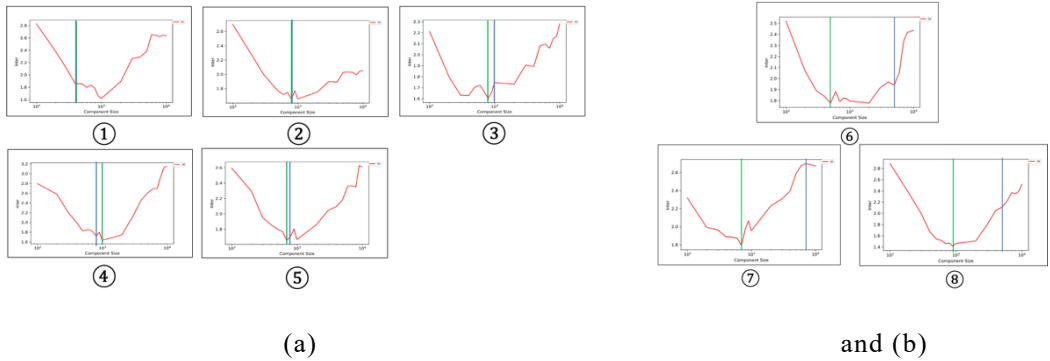

Figure 17. Graph of change in domain interaction (a) for Group A (b) for Group B

6 Conclusion

In this study, we proposed a technique to find a meaningful structure from the difference in the degree of entropy change when the region size and the component size were changed, by using hierarchical entropy and domain interaction to explain changes in the structure of an image. Considering Vermeer's girl with pearl earring as an example, we assumed that meaningful structures were found when the entropy was high. In fact, scientific research has shown that there is a hidden structure. In this way, it was shown that hierarchical entropy could lead to the discovery of a structure. For domain interaction, we hypothesized that the best balance of the segmentation of clusters was achieved at the component size when the domain interaction was at its minimum value. To test this hypothesis, we conducted a survey using a questionnaire. It was found that both the component size at which humans judged the best segmentation and the component size at which the domain interaction reached its minimum value coincided.

In the future, this study can be conducted in two ways. First is by using the method for segmentation of clusters [25]. The EGBIS used in this study was based on the RGB values of pixels only. If the texture of the objects in the image can also be treated as data, clustering of white clothes and walls will be possible, and if the component size is too small, many clusters become mosaic-like. If the component size is too large, the clusters are not divided. Therefore, a method to solve these problems is required. Other suitable segmentation methods need to be considered too.

Second, because capturing structural changes from hierarchical entropy is subjective, it is necessary to define a clear objective index. In previous studies dealing with earthquakes, entropy change has been used as an objective index to capture signs of seismic activity. However, in the field of imaging, it is necessary to discuss the meaning of the changes in the structure captured by entropy change. Through discussions, the meaning of the index will be brushed up, and the magnitude of the entropy change will become a widely used factor in determining which structures are important. In this way, the index could be useful in the discovery of hidden structures, as seen in the case of the girl with a pearl earring.